\documentclass[10pt,twocolumn,letterpaper]{article}

\usepackage[pagenumbers]{cvpr} %

\usepackage{graphicx}
\usepackage{amsmath}
\usepackage{amssymb}
\usepackage{booktabs}
\usepackage{makecell} %

\usepackage[pagebackref,breaklinks,colorlinks]{hyperref}

\usepackage{amsfonts}       %
\usepackage{nicefrac}       %
\usepackage{microtype}      %
\usepackage{xcolor}         %
\usepackage{graphicx}
\usepackage{colortbl}

\usepackage[capitalize]{cleveref}
\crefname{section}{Sec.}{Secs.}
\Crefname{section}{Section}{Sections}
\Crefname{table}{Table}{Tables}
\crefname{table}{Tab.}{Tabs.}

\renewcommand{\paragraph}[1]{\vspace{1.25mm}\noindent\textbf{#1}}
\newlength\savewidth

\newcommand{\tablestyle}[2]{\setlength{\tabcolsep}{#1}\renewcommand{\arraystretch}{#2}\centering\footnotesize}
\newcolumntype{x}[1]{>{\centering\arraybackslash}p{#1pt}}
\newcolumntype{y}[1]{>{\raggedright\arraybackslash}p{#1pt}}
\newcolumntype{z}[1]{>{\raggedleft\arraybackslash}p{#1pt}}

\definecolor{lightblue}{HTML}{0071BC}

\hypersetup{colorlinks,allcolors=lightblue, urlcolor=lightblue, citecolor=lightblue}

\definecolor{cvprblue}{rgb}{0.21,0.49,0.74}

\def\paperID{16284} %
\def\confName{CVPR}
\def\confYear{2025}

\newcommand{\ourshort}{\texttt{HMA}\xspace}
 
\begin{document}

\title{Learning Real-World Action-Video Dynamics with\\Heterogeneous Masked Autoregression}

\author{Lirui Wang$^1$ \quad Kevin Zhao$^{1*}$ \quad Chaoqi Liu$^{2*}$  \quad Xinlei Chen$^{3}$\\[1mm] $^1$MIT \quad $^2$UIUC \quad $^3$Meta, FAIR\\[2mm]\small{\href{https://liruiw.github.io/hma}{ \texttt{https://liruiw.github.io/hma}}} }

\maketitle

\begin{abstract}
We propose Heterogeneous Masked Autoregression (\ourshort) for modeling action-video dynamics to generate high-quality data and evaluation in scaling robot learning. Building interactive video world models and policies for robotics is difficult due to the challenge of handling diverse settings while maintaining computational efficiency to run in real time. \ourshort uses heterogeneous pre-training from observations and action sequences across different robotic embodiments, domains, and tasks. \ourshort uses masked autoregression to generate quantized or soft tokens for video predictions. \ourshort achieves better visual fidelity and controllability than the previous robotic video generation models with 15$\times$ faster speed in the real world. After post-training, this model can be used as a video simulator from low-level action inputs for evaluating policies and generating synthetic data.

\end{abstract}

\section{Introduction}
\label{sec:intro}
Scaling robot learning is bottlenecked by i) large amounts of high-quality data for training; ii) real-time and high-fidelity evaluation. This is in stark contrast with other domains such as natural language processing \cite{radford2019language,openai2023gpt4} and computer vision \cite{kirillov2023segment,videoworldsimulators2024} where abundant training data are available on the Internet and evaluation can be simply performed online with held-out data.

One compelling solution to these bottlenecks is generative modeling \cite{videoworldsimulators2024,1X_Technologies_1X_World_Model_2024}, which aims to learn the full dynamics and simulate the world. For training, it can produce an infinite amount of in-distribution data; for evaluation, it can potentially generate reasonable physical interactions without real-world deployment. However, building useful generative models is nontrivial. It has to be \emph{general} to handle various setups across embodiments, domains, and tasks rather than specific settings \cite{bruce2024genie,valevski2024diffusionmodelsrealtimegame,alonso2024diffusion}; it's also desirable to be \emph{efficient}, so as to deal with the real-time interactions from the policies~\cite{zhu2024irasim,yang2023learning} necessary for robotic applications.

\begin{figure}
\centering
\includegraphics[width=\linewidth]{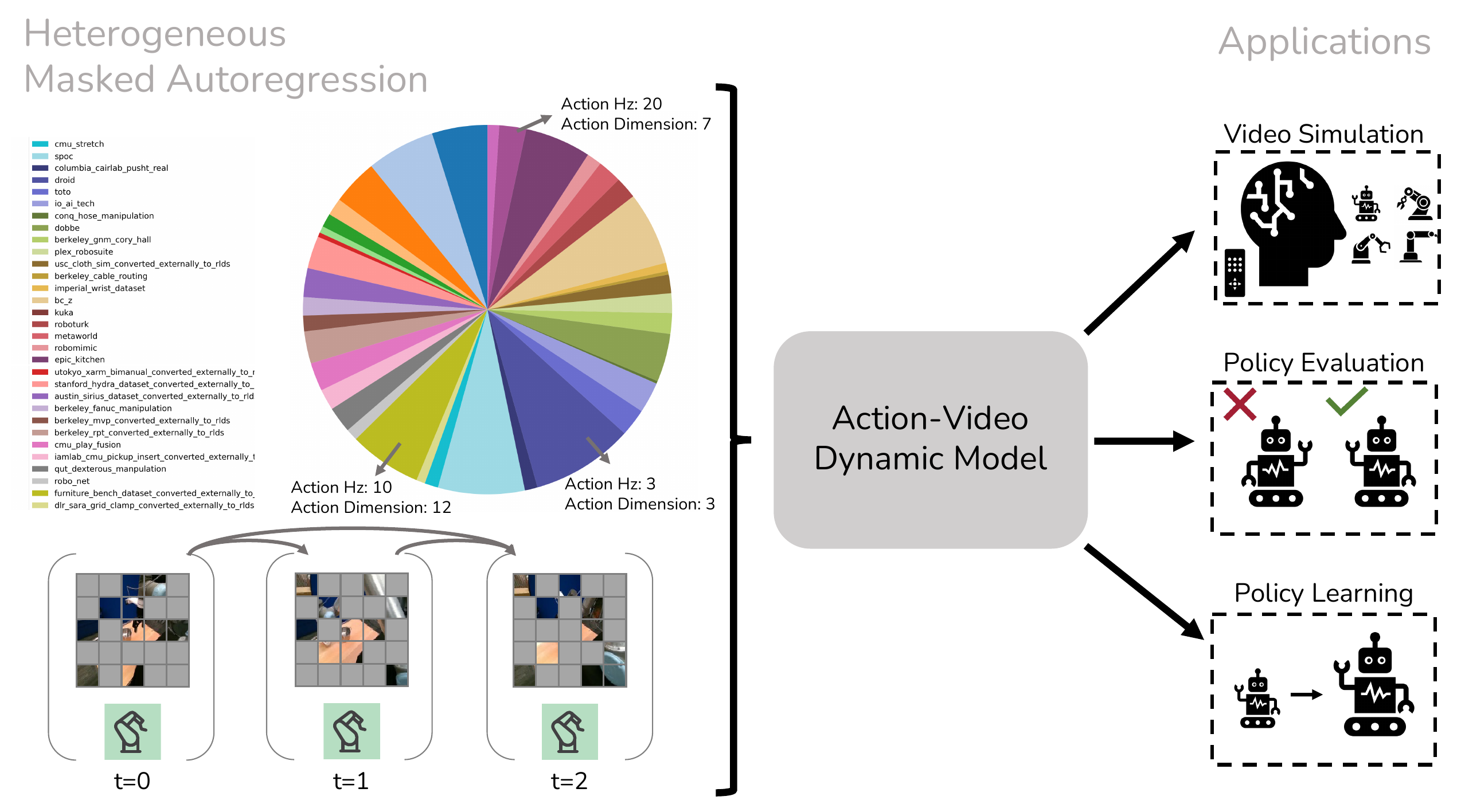}
\caption{\label{fig:frame}
\textbf{Action-Video Dynamics Model from Heterogeneous Robot Interactions.} \ourshort utilizes heterogeneous datasets comprising over 3 million trajectories (videos) from 40 distinct embodiments to pre-train a full dynamics model with next-set-of-token predictions using masked autoregression. After pre-training, the resulting action-video dynamics model is versatile, supporting applications such as video simulation, policy evaluation, synthetic data generation, and direct adoption as an imitation policy.}
\end{figure}

Building a general solution across embodiments, domains, and tasks at scale requires handling the \textit{action heterogeneity} in robotics. In particular, different robots use different action spaces, action frequencies, and action horizons for their specific tasks. 
To this end, we build on the recent idea that aligns such heterogeneity into the shared latent space \cite{wang2024scaling,doshi2024scaling,team2024octo} of action-video dynamics generation. To maximize the generality of the framework (\cref{fig:frame}), the network is modularized such that after pre-training, any new embodiment only requires training a small action encoder (``stem'') and action decoder (``head'') from scratch.

Beyond pure policy learning~\cite{wang2024scaling}, our full-dynamics setting also requires modeling of observations, which are typically formatted as sequences of images, or videos. While state-of-the-art, diffusion-based video modeling \cite{zhu2024irasim,videoworldsimulators2024,rigter2024avid} has achieved impressive visual quality, such frameworks are inefficient for real-time applications like robotics due to their need for extensive iterations over the entire sequence at each generation step. In contrast, we build on recent advances in autoregressive modeling for vision \cite{li2024autoregressive,liu2024mardini,tian2024visual}, which offer a more efficient alternative while maintaining high generation quality. In particular, we leverage masked autoregression \cite{chang2022maskgit,li2024autoregressive} for action-video dynamics and explore two variants to trade off speed \vs fidelity for videos: the discrete variant which generates vector-quantized (VQ) tokens at a high speed, and the continuous one which better preserves visual fidelity.

With these insights on Heterogeneity (\texttt{H}) on actions and Masked Autoregression (\texttt{MA}) on video dynamics, we introduce \ourshort, a masked autoregression framework for \textit{action-video dynamics} (AVD) over heterogeneous robotic data. The versatility of this unmasking architecture handles several core problems in robotics, including world models and end-to-end policies (\cref{fig:dynamics}). These problems can be framed as dynamic models using sequence modeling, enabling the joint generation of observations and actions \cite{chen2021decision, radosavovic2024humanoid}.

\ourshort is shown to scale across embodiments, trajectories, and model sizes in the heterogeneous pre-training mixtures. In particular, we study the scaling behaviors measured by visual fidelity and action controllability. The model is pre-trained with over 3 million trajectories (videos) with action labels and a single \ourshort can generate video across a wide range of 40 embodiment datasets from 2-DoF action space to 28-DoF action space.

The main use case of \ourshort is the first high-fidelity and real-time video robotic simulator (\href{https://liruiw.github.io/hma/hma_demo}{example}). Our generation achieves better visual fidelity and controllability than the previous state-of-the-art \cite{zhu2024irasim} with \textbf{15$\times$ faster} inference latency, enabling real-time interactions \cite{lynch2023interactive}. Because \ourshort autoregressively models the video and action sequences in interaction data, it can also be used to generate trajectories over 100 frames (over 30 seconds) stably in various robotic applications. On simulation benchmarks \cite{robomimic2021}, \ourshort is used to evaluate policies as a high-fidelity simulator and to generate 90\% synthetic trajectory data for improving policy performance to match ground truth data. The learned dynamic models can also be used as an imitation policy to predict actions. We hope that \ourshort can shed
some light on learning action-video dynamic models in a unified framework from heterogeneous data.

\section{Related Works}
\label{sec:related_works}

\paragraph{World Models.}
 World models \cite{ha2018world}, or dynamic models \cite{bertsekas1995neuro}, are computer programs that evolve based on agents' behaviors. Different from simulator software, learned dynamic models predict future states or reward functions based on past observations and then apply to model-based reinforcement settings \cite{zhang2021autoregressive,chen2021decision,hansen2023td} or robotics \cite{byravan2017se3,li2018learning,seo2023masked}. Notably, since ground truth states are often unavailable in decision-making, dynamic models need to handle high-dimensional video data and low-dimensional grounded physical actions. Full-sequence diffusion and autoregressive models are two primary approaches for generative tasks across languages, images, and videos~\cite{li2024autoregressive,ye2024latent,kondratyuk2023videopoet,chang2022maskgit,ho2022video}. In particular, 1xGPT~\cite{1X_Technologies_1X_World_Model_2024} uses masked autoregression for video generation and MAR~\cite{li2024autoregressive} applies diffusion losses with masked autoregression for image generation, and Diffuser \cite{janner2022planning} uses diffusion models to jointly model the full state and action sequences in planning. Our work focuses on masked autoregression over full-sequence diffusion \cite{bruce2024genie,valevski2024diffusionmodelsrealtimegame,alonso2024diffusion}, aiming to create efficient and interactive world models.
 
\paragraph{Steering Video Models with Actions.}
Video models \cite{videoworldsimulators2024} can be applied to a wide range of video data including curated videos, human videos, synthetic videos, and robot videos. Interactive video models usually rely on language instructions \cite{yang2023learning}, latent actions \cite{bruce2024genie}, or sketches to guide the video or image models.  However, video controllability is still an issue when \textbf{fine-grained details and low-level motion controls} are  used as prompts \cite{zhu2024irasim,rigter2024avid,alonso2024diffusion}. In particular, IRASim \cite{zhu2024irasim} applies DiT \cite{peebles2023scalable} to several robotic datasets and demonstrates high-quality video simulation qualities. Unlike previous works that apply such models to a single task or embodiment or use natural languages as actions, our work investigates action-conditioned video models across heterogeneous embodied settings and their scaling behaviors.  

\paragraph{Visual Generative Models for Robotics.}
Visual generative models have been explored extensively for robotic applications such as policy learning, planning, and synthetic data generation. Synthesized goals and subgoals \cite{du2023video,nair2018visual} have been used to improve end-to-end manipulation policies. Video predictions such as visual foresight \cite{finn2017deep} have been used to guide policy executions. Video language planning \cite{du2023video} achieves long-horizon planning tasks through model-predictive control and search. In the context of robotics, diffusion methods have been used to augment images 
 \cite{chen2023genaug,yu2023scaling} and 3D generative methods have been used to generate synthetic data from novel views \cite{zhou2023nerf}. In this work, we use learned dynamic models for policy evaluation and synthetic data generation in policy learning.  Moreover, the dynamics model can act as policies for action predictions.

\section{Heterogeneous Masked Autoregression}
\label{sec:method}
\begin{figure}
    \centering    \includegraphics[width=\linewidth]{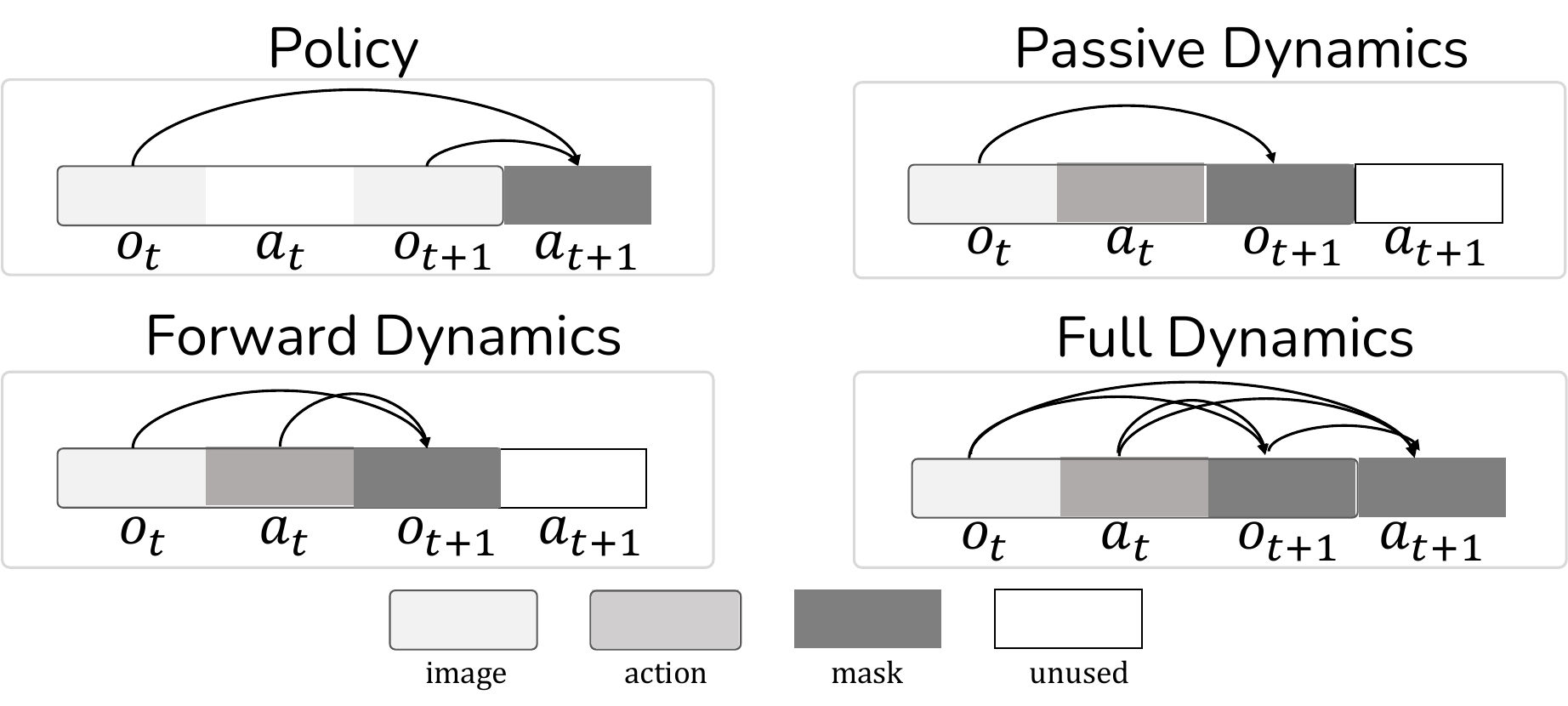}
    \caption{\textbf{Dynamics Model.} Masked autoregression in the dynamics model generalizes multiple problem settings including policy learning, forward and passive dynamics, and full dynamics. }
    \label{fig:dynamics}
\end{figure}

\subsection{Dynamic Models}
\label{subsec:dynamics}
Our objective is to learn the dynamics model $f$ on visual observation history $\mathcal{O}_{\text{history}}=\{o_{t-N_{\text{past}}},...,o_{t-1}\}$ and action history $\mathcal{A}_{\text{history}}=\{a_{t-N_{\text{past}}},...,a_{t-1}\}$ to predict the future observations $\mathcal{O}_{\text{future}}=\{o_{t},...,o_{t+N_{\text{future}}}\}$  and actions $\mathcal{A}_{\text{future}}=\{a_{t},...,a_{t+N_{\text{future}}}\}$ where $N_{\text{past}}$ and $N_{\text{future}}$ are hyper-parameters. Using the masking objective in the next section, the model is trained to predict a masked component in this function given other components. In \cref{fig:dynamics}, multiple core problems in decision-making and robotics \cite{bertsekas1995neuro,murray2017mathematical} can be viewed as a subset of this problem: 
full dynamics: $f_{\text{full-dynamics}}  (\mathcal{O}_{\text{history}},\mathcal{A}_{\text{history}})\mapsto (\mathcal{O}_{\text{future}},\mathcal{A}_{\text{future}})$, passive video predictions $f_{\text{passive}}  (\mathcal{O}_{\text{history}})\mapsto \mathcal{O}_{\text{future}}$, 
forward dynamics: $f_{\text{forward-dynamics}}(\mathcal{O}_{\text{history}},\mathcal{A}_{\text{history}})\mapsto \mathcal{O}_{\text{future}}$,  and policy models: $f_{\text{policy}}(\mathcal{O}_{\text{history}},\mathcal{A}_{\text{history}})\mapsto \mathcal{A}_{\text{future}}$ are all subsets of such sequential and generative models. The objective can be applied to datasets with missing labels such as pure video datasets and policy datasets, and can even be extended to inverse dynamics $f_{\text{inverse-dynamics}}  (\mathcal{O}_{\text{history}},\mathcal{O}_{\text{future}})\mapsto \mathcal{A}_{\text{future}}$ if non-causal models are used.

\subsection{Action Heterogeneity}
A world model that understands the physical world should be able to simulate various forms of embodiments and actions. For instance, humans can intuitively understand how other creatures' physical actions would impact the world. Action heterogeneity in robotics includes different action spaces, action frequencies, and action dimensions. For example, a dynamics model with 30Hz frequencies and a 50-DoF joint space for a humanoid is different from one with 5Hz frequencies and a 6-DoF end effector space for a Franka Arm. Since the videos can be unified as a sequence of images of fixed resolutions, we decouple observation and actions to handle the heterogeneity in actions. One particular approach that handles such action heterogeneity without explicit tokenization processing step is HPT~\cite{wang2024scaling}. Specifically, for each domain, we consider 2Hz dynamics frequencies and 12 frame context (6 seconds wallclock time) to balance generation length and compute efficiency, but technically any action frequencies can be used. We chunk the action sequences into a fixed 2Hz to ensure consistency across datasets. For action prediction objectives, we will use different decoder heads to predict actions for different embodiments.

\subsection{Masked Autoregression}
Following the success of language models and image/video models, we use masked autoregression (MAR) \cite{1X_Technologies_1X_World_Model_2024,chang2022maskgit} to generate video predictions and action predictions based on previous video observations and action sequences. Specifically, masked autoregression models the joint distribution as the conditional distribution based on previous generations. It uses the masked autoencoding objective with a random order of tokens at training time. Then it predicts the next-set-of-tokens and
gradually unmasks at inference time \cite{li2024autoregressive}, which is natural in robotic interactions. The joint dynamics probability distribution can be decomposed as:

\begin{align}
p(\mathcal{O}_{\text{history}},\mathcal{O}_{\text{future}},\mathcal{A}_{\text{history}},\mathcal{A}_{\text{future}})=p(X_1,...,X_K) \nonumber \\
   =\Pi_{k=1}^K p(X_k|X_1,...,X_{k-1}),
\end{align}
where $X_k$ can be any (causally) valid masked set of observations and actions. Therefore, it can unify the different dynamic settings mentioned in \cref{subsec:dynamics}. We use a neural network $f_\theta$ parameterized by $\theta$ to learn such a distribution.  Notably, both images and actions are continuous signals, and multiple loss functions can be applied to this. One simple choice is to tokenize the images into discrete tokens and use the raw actions. We then train the network with cross-entropy (CE) loss on image tokens while applying mean-squared-error (MSE) regression losses to the actions. For simplicity, let ${X}=(\hat{o}, \hat{a})$ generated by $\theta$ and $({o}, {a})$ denote the ground truth, the discrete loss on the video vector-quantized (VQ) tokens can be written as:
\begin{equation}
\label{eq:discrete}
    \mathcal{L}_{\text{VQ}}(X;\theta)=MSE(a,\hat{a})+CE(o,\hat{o}).
\end{equation}

Alternatively, we can use a \textbf{separate} denoising diffusion objective that learns to reconstruct $X$ based on a continuous latent $z$, which we also call ``soft tokens'' ~\cite{li2024autoregressive}: 
\begin{equation}
\label{eq:continuous}
    \mathcal{L}_{\text{soft}}(X;\theta)=\lVert \epsilon_a - f(a|t,z) \rVert^2  +\lVert \epsilon_o - f(o|t,z) \rVert^2,
\end{equation}
where $\epsilon_a,\epsilon_o$ are noise vectors sampled from $\mathcal{N}(0,\mathbf{I})$ and $t$ is the timestep of the noise schedule. Note that $z,t$ are separate for action $a$ and video $o$ in practice.

Different from previous video models that use either diffusion \cite{zhu2024irasim,alonso2024diffusion} or autoregression \cite{liu2024mardini,bruce2024genie}, this method combines the efficiency and expressiveness of both approaches to model the joint dynamic models.

\subsection{Model Architecture}
The overall architecture, denoted as  Heterogeneous Masked Autoregression (\ourshort), builds on the generality across embodiments with action heterogeneity, and the efficiency across dynamic settings and discrete/continuous observation tokens with masked autoregression.

\begin{figure}
    \centering
    \includegraphics[width=\linewidth]{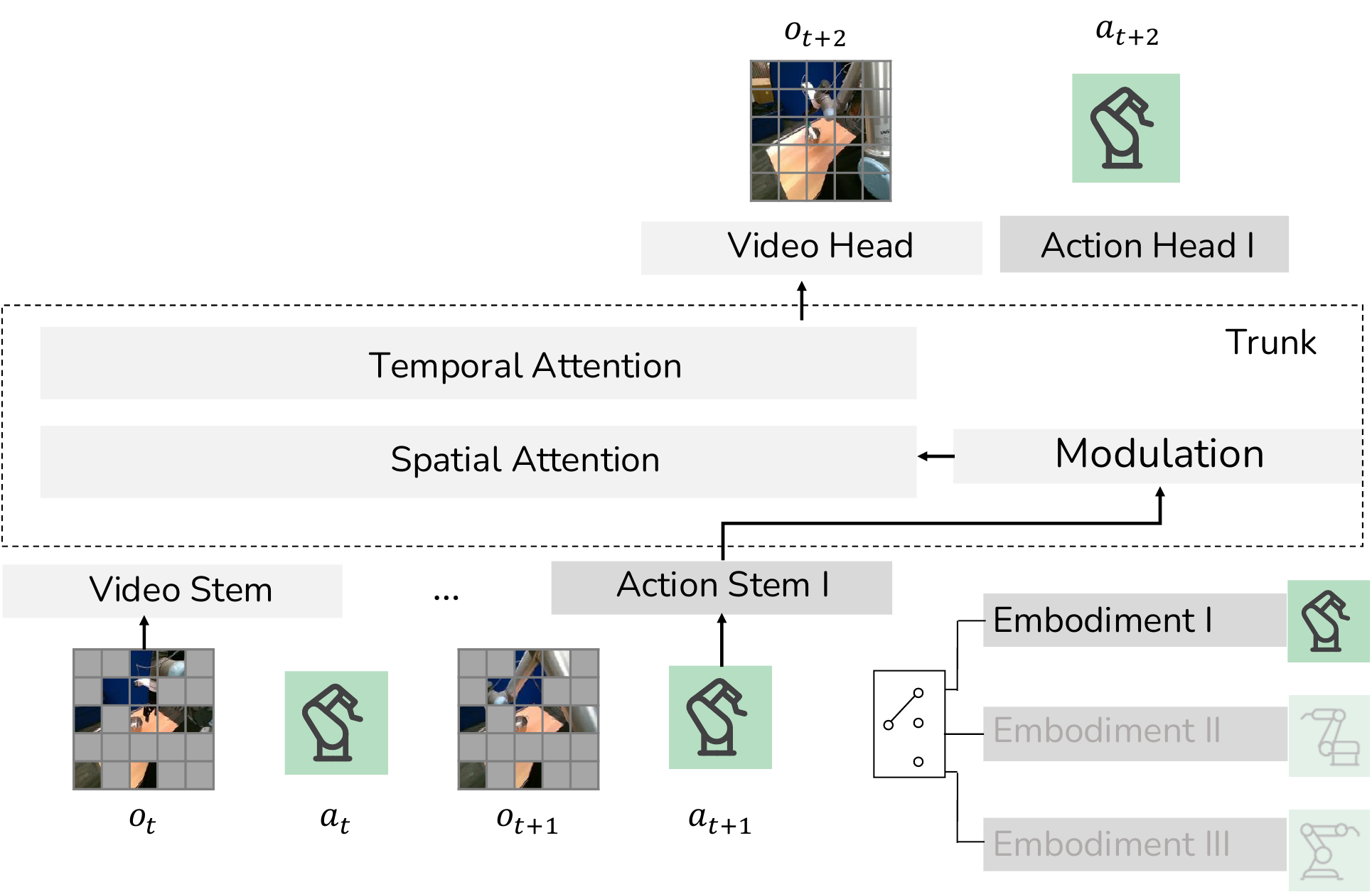}
    \caption{\textbf{Network Architecture.} The \ourshort model architecture maps low-level video and action sequences across different embodiments into a shared latent space. For actions, embodiment projectors are activated based on the training sample. The spatial-temporal Transformer produces the output video and action tokens for future frames.   }
    \label{fig:network}
\end{figure}

In \cref{fig:network}, the network architecture of \ourshort follows the heterogeneous pre-training in HPT \cite{wang2024scaling}, in which we create multiple modules of action inputs (``stem'') and action outputs (``head'') and share the core spatial-temporal transformer (``trunk'') as the dynamic models for pre-training and transferring. The spatial attention runs bi-directionally with the masked video and action tokens and the unmasked tokens, and the temporal attention is causal in predicting tokens in future steps.

This model architecture explicitly handles the heterogeneity in action spaces in robotics \cite{wang2024scaling,shazeer2017outrageously}. Specifically, each domain has its own action encoder which has an MLP layer that maps normalized actions of certain horizons into features. A decoder can be a 3-layer diffusion MLP that is trained to regress on the continuous features and/or actions.

Notably, the ability to predict a controllable and high-fidelity future given past observations requires dedicated information streams from the actions. We use modulation \cite{peebles2023scalable} layers for each domain in every Transformer block \cite{shazeer2017outrageously}. To generalize the video model to predict both video and action in the full dynamics task, we also use the token concatenation method for fusing video and action tokens.

\subsection{Training and Inference}

In practice, each frame in the video has 256 tokens and 64 repeated action tokens. The masking has a minimum ratio and follows a cosine schedule to mask more tokens at later steps in the training horizon. We  use Maximal Update Parametrization \cite{yang2022tensor} for scaling with bigger models.  We find a Xavier initialization with gain 0.1 for the action projectors to be useful for training stability.  We use different mask tokens and different decoders for action and video tokens. We ablate with other variants such as cross-attention layers in the Section \ref{sec:experiment}.

At inference time, we append masked tokens to the video sequence and masked tokens to the action sequence up to the maximum horizon whenever needed. The full inference process for the diffusion-based \ourshort contains three nested autoregression procedures across $T$ timesteps in the video time dimension, $M$ unmasking steps in the image patch dimension, and  $N$ diffusion steps for the continuous token generation. Since each step in the nested iterations is fast, the generation process is both high-quality and efficient. For diffusion head, we use DDIM with per-step clipping \cite{song2020denoising} to train with $N_{\text{train}}=1000$ timesteps and test with $N_{\text{test}}=100$ steps. As in standard diffusion works \cite{peebles2023scalable}, we use patch size 2 to reduce context length. This means a $2\times2$ ``patch" of tokens is replaced with a single token, We find $M=2$ unmasking \cite{chang2022maskgit,li2024autoregressive} iterations with random unmasking order  to be sufficient to generate high-quality videos at inference time and iterations across $T$ are not necessary.

\section{Pre-Training Experiments\label{sec:experiment}}

We first discuss the implementation details below for training \ourshort, including datasets, models, and evaluation metrics.
\begin{figure*}[t]
\centering
\includegraphics[width=.95\linewidth]{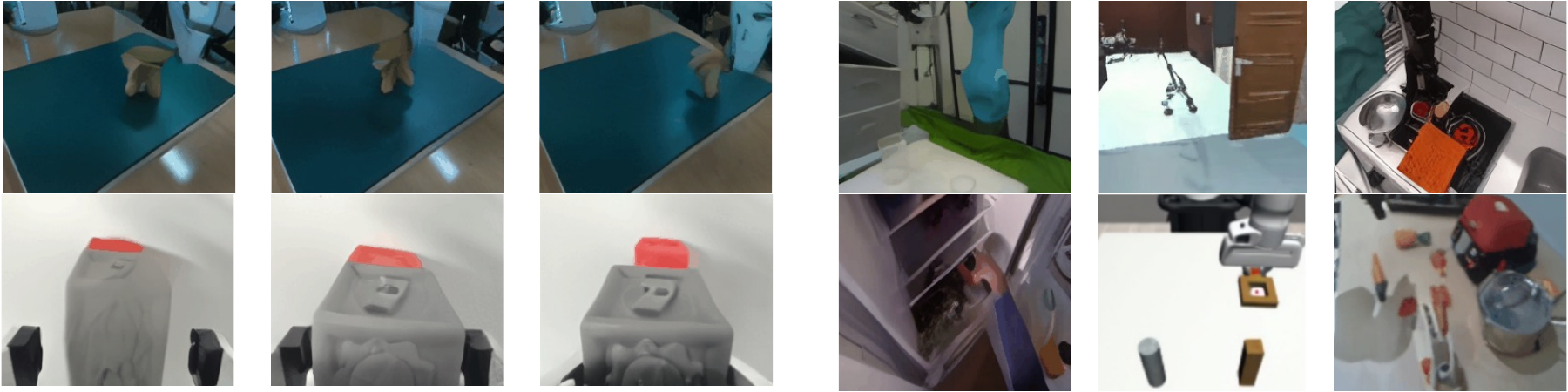}
\caption{\label{fig:video_results}\textbf{Pre-trained Video Model Generation.} We show that a single unified \ourshort model can generate realistic (left 3 columns) and diverse (right 3 columns) videos across multiple embodiment datasets with heterogeneous action spaces. Each group shows three generated frames from a single sequence.}
\end{figure*}

\begin{figure}
    \centering    \includegraphics[width=\linewidth]{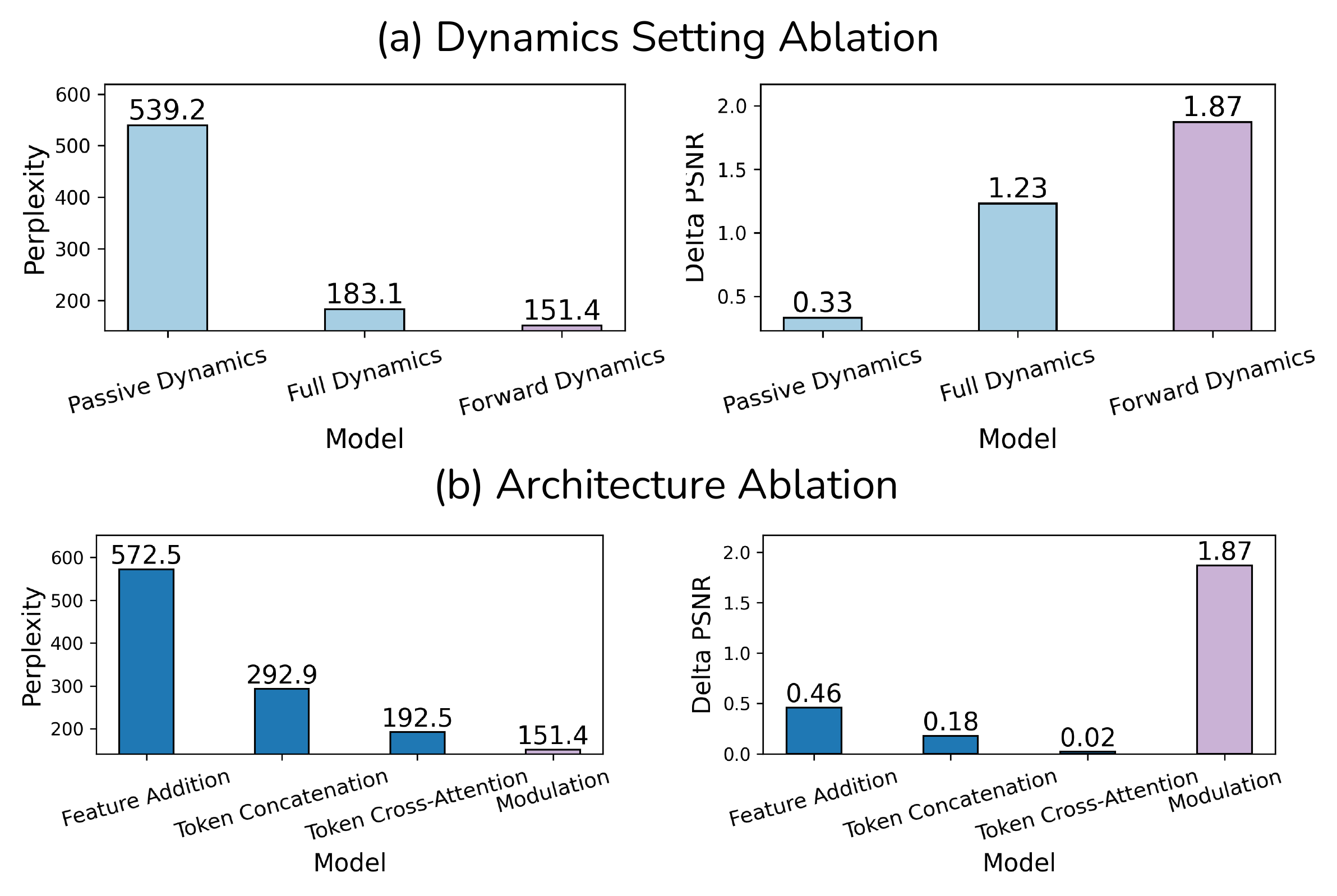}
    \caption{\textbf{Ablation on Pre-training Settings and  Architecture.} Under the pre-training setting with VQ tokens, we ablate the video generation performance (visual fidelity measured by perplexity and controllability measured by controllability). (a) We find action-conditioned models outperform passive video models. (b) We compare different action conditioning architectures in the masked autoregression framework. The purple color denotes the best model that we use by default. }
    \label{fig:pretrain_ablation}
\end{figure}

\paragraph{Datasets.} We use multiple domains of embodiment data related to robotic manipulations. In the largest dataset mixture, the 35 real robot datasets contain primarily the Open-X embodiment dataset \cite{o2023open}, and 3 human video datasets contain ego-centric hand motions \cite{grauman2024ego,damen2020epic,damen2022rescaling}  and the 2 simulation datasets contain standard benchmark \cite{robomimic2021,yu2020meta}. We follow the same dataset processing as in previous works \cite{wang2024scaling} for using 2D hand detections as action label proxies.  The full dataset has over 2.5 billion frames and 3 million trajectories. Note that these datasets with action labels usually have different action dimensions and action frequencies.  During training, we apply inverse exponential probability for sampling the dataset of each batch sample. We preprocess videos in these datasets into tokens with fixed resolution ($256\times 256$) using the pre-trained Stable Video Diffusion VAE \cite{blattmann2023stable} with $8\times8$ spatial downsampling for implicit diffusion models and the 1XGPT tokenizer \cite{1X_Technologies_1X_World_Model_2024}, which is a fine-tuned OPEN-MAGVIT2 \cite{luo2024open} tokenizer, with $16\times16$ downsampling for explicit cross-entropy losses. We unify all datasets into 2Hz by adaptively choosing action strides.  For example, a 10Hz dataset will have an action stride of 5 and shrinking into 2Hz action chunks. Under 2Hz frequencies, the context length is in total 12 frames (6 seconds) with 4 frames (2 seconds) as prompts and 8 frames (4 seconds) as predictions, which are sufficient for most closed-loop interactive applications. We choose these hyperparameters for prototyping, but technically any image resolution or video frequencies can be used. The continuous objective also does not require a tokenizer and can directly operate on pixel space.

\paragraph{Model.} Under the standard setup for ablations,  we use the discrete model and VQ tokens with forward dynamics objective (with cross-entropy loss) for the simplicity of training. We use a 32-layer transformer model with dimension $d=256$. The model is trained with 8 V-100 GPUs and batch size 64 for around 60,000 iterations and 2 epochs. We train larger models with up to 64 GPUs.

\paragraph{Metrics.} We measure the performance of the video models via several metrics on the held-out validation datasets with a maximum of 500 examples per dataset and present the average statistics across datasets. When we use auto-regressive models with explicit cross-entropy loss on tokenized outputs, we measure the training objectives for \textit{fidelity} via validation perplexity, which is directly correlated with validation loss and visual fidelity metrics, such as PSNR, SSIM. We use $\Delta$PSNR \cite{bruce2024genie} to measure \textit{controllability}, which is the average difference of PSNR on the last frame computed from ground truth action sequence with PSNR computed from 5 random action sequences. Intuitively, if $\Delta$PSNR is small, then the model predictions are less affected by actions and therefore the video model is not controllable.

\begin{figure*}
    \centering
    \includegraphics[width=\linewidth]{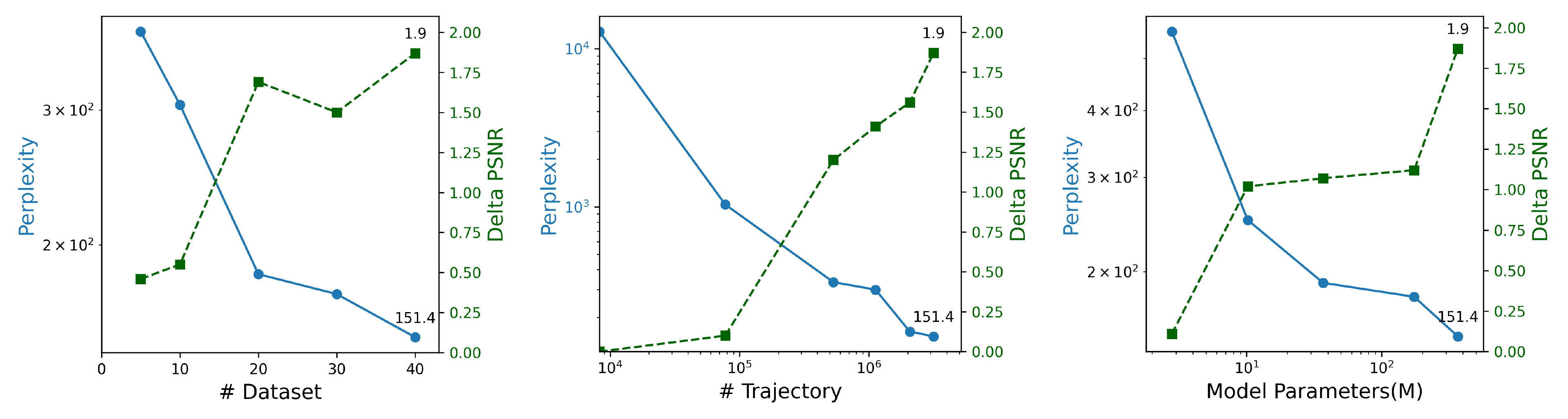}
    \caption{\textbf{Experiments on Scaling Behaviors of \ourshort.} We observe positive trends in the scaling performance of heterogeneous video models across axes including the number of datasets, number of trajectories, and model sizes. The evaluation metrics on fidelity (perplexity) and controllability ($\Delta$PSNR) are averaged across validation datasets.  }
    \label{fig:pretrain_scaling}
\end{figure*}

\subsection{Action-Video Dynamics under Heterogeneity\label{sec:hma_pretrain}}

In this experiment section, we investigate how well heterogeneous pre-training works for world models. We choose the discrete loss objective in \ourshort for these experiments. In \cref{fig:video_results}, we showcase the generation of our single unified pre-train model \ourshort to handle heterogeneous datasets.

\paragraph{Dynamic Model Settings.} In this section, we measure the performance of different dynamics model settings, including passive video generation, forward dynamics (action-conditioned video generation), and full dynamics which have an auxiliary task of action predictions. Shown in \cref{fig:pretrain_ablation} (a), low-level action conditioning in the forward dynamics model can improve video generation results. Intuitively, passive video sequences without actions contain less information on the causal physical relationships of these videos and are sometimes hard (or even ambiguous) to capture fine-grained details related to motions. We find full dynamics training outperforms passive dynamics but does not outperform forward dynamics generation. We hypothesize that these results stem from the limitations of VQ tokens, which were originally designed for visual generation tasks rather than for action prediction.

\paragraph{Action Architecture Ablation.} We compare several architectural designs in conditioning action information for video generation. Specifically, we ablate architecture variants including i) modulation, ii) token concatenation, iii) feature addition, and iv) token cross-attention following \cite{peebles2023scalable}. In \cref{fig:pretrain_ablation} (b), we show that the modulation method can outperform other methods.  In particular, token concatenation along the sequence dimension for each frame does not have enough expressiveness and compute on action-conditioning, compared to per-layer modulation.

\subsection{Scaling Behaviors of \ourshort}
We experiment with the scaling performance of \ourshort along multiple axis. In \cref{fig:pretrain_scaling}, we scale across the number of videos per dataset, scale across the number of datasets, and scale across model sizes. Under the same validation datasets, we observe consistent gains from scaling measured by perplexity and $\Delta$PSNR.  

\paragraph{Scaling Embodiments.} Scaling the number of datasets (\cref{fig:pretrain_scaling} Left) from 5 to 40 datasets strictly increases the dataset heterogeneity and creates more model parameters to handle such action space differences among embodiments. Yet we see that the video prediction performance does not degrade or become unstable. This gives positive signals for joint heterogeneous training with diverse embodiment video datasets. For this experiment, we evaluate the models on the first random subset of 5 datasets to be consistent. 

\paragraph{Scaling Data.} Using 40 datasets, we scale the number of trajectories (videos) per dataset from 10 to $10^6$, equivalent to training on 8 thousand to 3 million trajectories. In  \cref{fig:pretrain_scaling} middle, the slight plateau of the trajectory scaling performance after $10^5$ total trajectories is likely due to the dataset quantity imbalance, as the additional trajectories all come from the largest few datasets. 

\paragraph{Scaling Model.} Under the largest dataset with the maximum number of datasets (40) and maximum trajectories (3 million), we study model scaling of \ourshort. In  \cref{fig:pretrain_scaling} right, we scale the hidden dimension of the Transformer while keeping other parameters constant. This changes the model parameters from 3 million to 400 million (sparse) parameters. We qualitatively observe improved visual qualities and controllability when we increase the model capacity.

\section{Post-Training Applications\label{sec:post}}

After pre-training, we finetune \ourshort to evaluate its performance with limited data and ground truth controllability metrics. In the real-robot action simulator setting,  we demonstrate its speed advantages over the previous state-of-the-art interactive simulator by 15$\times$. We also post-train the model to various downstream robotics use cases including policy evaluation, policy data generation, and dynamic models as policies.

\paragraph{Datasets.} We focus on Robomimic \cite{robomimic2021}, a simulation benchmark with 200 trajectories per task, and Language Table \cite{lynch2023interactive}, a real-world benchmark with 442k trajectories, to study the applications of \ourshort model in simulation and policy learning. Simulation benchmarks are suitable because we can evaluate the ground truth controllability and benchmark against policy methods and policy evaluation. Language Table is suitable for its sheer amounts of trajectory data and focused domain. We use held-out videos to evaluate \ourshort on these two datasets. Since we need to apply interactions in post-training, we use an action stride of 1 and maintain the original action frequencies in the dataset.

\paragraph{Model.} Following the training protocol of \cref{sec:experiment}, we finetune the pre-trained \ourshort with layer dimension 256 on each dataset for 10 epochs. The interactive simulation setting requires the forward dynamics formulation of our model. In \cref{tab:size_fps}, we compare different model sizes of \ourshort with the previous state-of-the-art open-source model IRASim \cite{zhu2024irasim} which uses the DiT \cite{peebles2023scalable} model and action modulation architecture and outperforms VDM and LVDM \cite{he2022latent,
ho2022video} in robotics. Notably, IRASim is not optimized for single-frame interactions, as the default setup predicts video sequences based on action sequences. On \cref{tab:size_fps}, we observe a 15$\times$ model speed improvement (4.44 v.s. 0.28 FPS) even compared to their amortized result, thanks to masked autoregression. Using the diffusion heads, the model can still be faster while maintaining high visual fidelity.

\paragraph{Metrics.} For real-world data, we use model-based metrics such as PSNR \cite{hore2010image} and SSIM \cite{wang2004image}, learning-based metrics such as LPIPS \cite{zhang2018perceptual}, FID \cite{heusel2017gans}, and FVD \cite{unterthiner2018towards} for visual fidelity measurement; and $\Delta$PSNR \cite{bruce2024genie} for controllability. For simulation tasks, since we can access ground truth observations in any state through rendering, we directly compare ground truth against the learned model to measure controllability rather than using proxy $\Delta$PSNR. In particular, we apply perturbations to the action sequences and denote the metrics as PSNR$^*$ and perplexity$^*$ which measure the average sensitivity of the dynamics model to small action perturbations.

\begin{table}[t]
\centering
\small
\tablestyle{5.5pt}{1.2}
\begin{tabular}{lccl}
\textbf{Model}       & \textbf{Method}       & \textbf{Parameters} (M) & \textbf{FPS $\uparrow$}   \\ 
\hline
IRASim-XL   & DiT &  679            & 0.28 \\
IRASim-XL, amortized  & DiT & 679            & 0.58  \\ 

\hline

\ourshort-Base  & MaskGIT &  44             &  22.72 \\
\ourshort-XL & MaskGIT &  679         & 4.38 \\

\ourshort-Base     & MAR &  96          &  4.44   \\
\ourshort-XL    & MAR & 741         &  2.01  \\
\hline
\end{tabular}
\caption{\textbf{Inference Speed.} We measure the per-frame inference speed across 16 frames for various model sizes. The Base model has a model size of around 30M and the XL model has a similar model size as IRASim-XL. The models all use 32-block transformers where the base model has dimensions 256 and the XL models have dimensions 768. Our fastest model of the same size is more than 15$\times$ faster than \cite{zhu2024irasim} because \ourshort does not pass through the full Transformer multiple times (with diffusion modeling) to generate each frame. MAR incurs more parameters than MaskGIT \cite{chang2022maskgit} because of the diffusion heads \cite{li2024autoregressive}.  The amortized result for \cite{zhu2024irasim} comes from averaging over multiple frames. The speeds are all measured on the same hardware setup with RTX-4080 GPU. }
\label{tab:size_fps}
\end{table}

\begin{table}[t]
\centering
\small
\tablestyle{4pt}{1.2}
\begin{tabular}{l|cccccc}
& {\bf PSNR $\uparrow$} & {\bf SSIM $\uparrow$} & {\bf $\Delta$PSNR $\uparrow$} & {\bf LPIPS $\downarrow$} & {\bf FID $\downarrow$} & {\bf FVD $\downarrow$} \\ 
\hline
IRASim & 25.41 & 0.82 & 5.78 & 0.08  & 23.22 & 152.20\\
\ourshort &28.19 & 0.83 & 6.06  &    0.07 & 33.56 & 111.52
\end{tabular}
\caption{\textbf{Comparison with IRASim.} In Language Table Benchmark \cite{lynch2023interactive}, we show that a pre-trained \ourshort-based model (diffusion) is able to achieve better visual qualities and controllability than IRASim while maintaining faster speed and requiring less compute. The results are computed over 200 held-out trajectories.}
\label{table:compare_withirasim}
\end{table}

\subsection{Toward Real-Time Simulation with \ourshort}
\label{exp:finetune_real}
\begin{figure}[t]
\centering
\includegraphics[width=\linewidth]{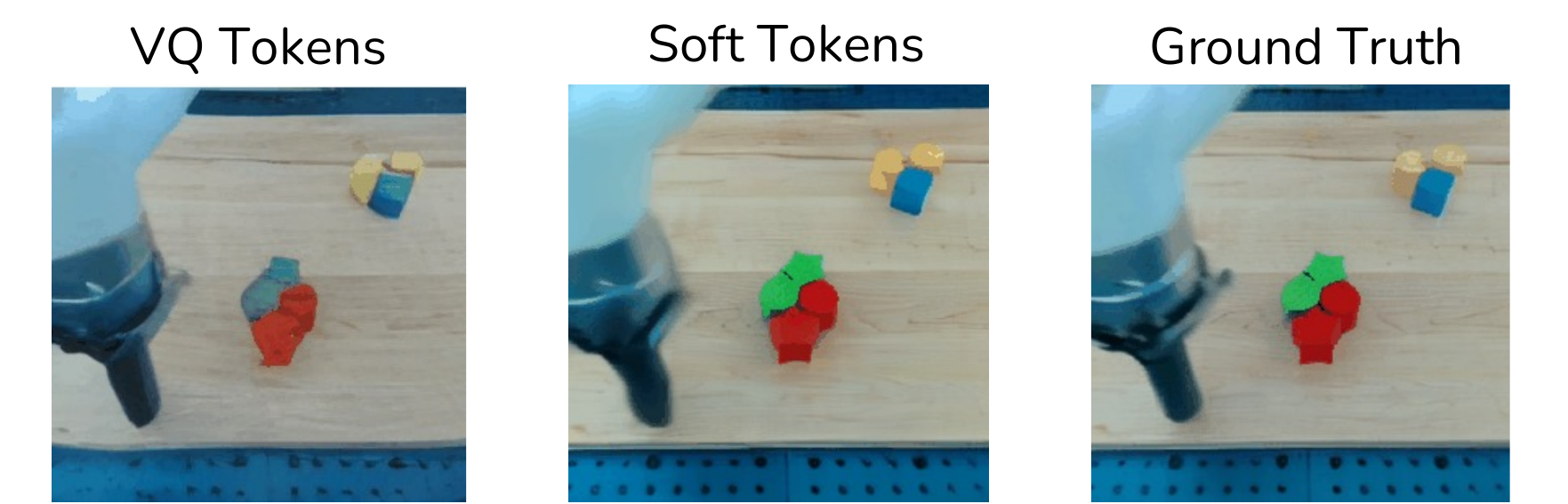}
\caption{\label{fig:tokenizer}\textbf{Qualitative Comparisons Between Tokenizers and Models.} Despite longer convergence time, diffusion-based methods (\cref{eq:continuous}) on soft tokens generate better visual quality than on VQ tokens (\cref{eq:discrete}), qualitatively and measured by PSNR.}
\vspace{-2mm}
\end{figure}

In this section, we illustrate how to use \ourshort as a real-time learned simulation model in robotics and compare it to IRASim \cite{zhu2024irasim} for controllability and visual fidelity.  First, in \cref{tab:finetune_real}, we observe better visual fidelity and controllability by using a pre-trained \ourshort model with discrete image tokens. In  \cref{fig:tokenizer}, we qualitatively compare different tokenizers along with training objectives. We find soft tokens to contain richer visual information in general, but take more space and compute to store and genereate.

\begin{figure}[t]
\centering
\includegraphics[width=\linewidth]{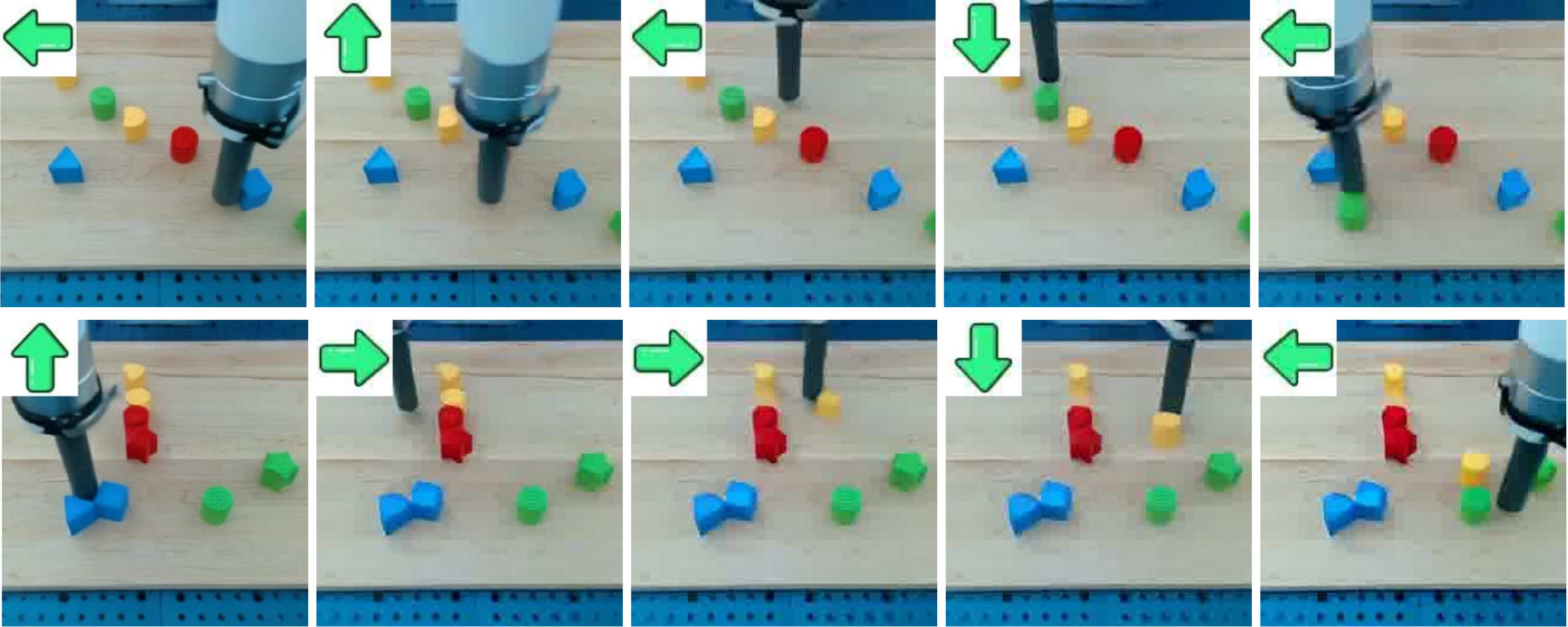}
\caption{\label{fig:controllability_comparison}\textbf{Video Controllability.}  \ourshort can follow user action inputs to generate physically plausible object permanence (top row) and block pushing interactions (bottom row). These video predictions are both at out-of-distribution settings and at a much longer horizon than training (over 100 frames). }
\vspace{-2mm}
\end{figure}

In \cref{table:compare_withirasim}, our MAR model achieves better visual fidelity and controllability compared to IRASim \cite{zhu2024irasim} despite using less than 1/8th of the parameters (\cref{tab:size_fps}).   When using a similar number of parameters, \ourshort is significantly faster than the amortized version of \cite{zhu2024irasim}. Our fastest discrete model can run up to 22Hz, which allows for real-time interactions. The speedup compared to full-sequence diffusion comes from the autoregression process and the architecture, which only requires computation on the diffusion heads rather than the full Transformer. Compared to full-sequence diffusion models, autoregressive models learn to predict one frame/action at a time, which is natural for robotic interactions. Qualitatively, we also find interacting with our learned simulators to be more reactive and consistent than \cite{zhu2024irasim}, with only small compounded errors in generation quality when running long autoregressive inferences (e.g. \cref{fig:controllability_comparison}) Given the scaling performance (\eg  \cref{fig:pretrain_scaling}) in \cref{sec:experiment}, we expect training longer with more compute resources to further improve the generation quality. 

In \cref{fig:controllability_comparison}, we build a simple user interface to test the performance of \ourshort under an out-of-distribution setting and find stable simulation quality. Under real-world compute and inference speed constraints for robotic policy and simulation use cases,  \ourshort with masked autoregression is more suitable than full-sequence diffusion, which can be of order faster to reach real-time speed. 
This showcases the potential of \ourshort for policy evaluation and synthetic data generation in the real world, which can save effort in evaluation and collecting data when scaling robot learning. 

\begin{table}[t]
\centering
\small
\tablestyle{7pt}{1.2}
\begin{tabular}{l|cccc}
& {\bf PSNR $\uparrow$} & {\bf Perplexity $\downarrow$} & {\bf $\Delta$ PSNR $\uparrow$} & {\bf LPIPS $\downarrow$}   \\ 
\hline
\ourshort  & 21.01 &  305.87 & 0.01 &  0.19 \\
\ourshort$^{+}$ &  22.04  & 189.83 & 0.06 &   0.17
\end{tabular}
\caption{\textbf{Real World Finetuning.}  \ourshort$^{+}$ denotes finetuned model based on pre-trained checkpoints while \ourshort trains from scratch on the finetuning data. This experiment uses the discrete loss baseline. }
\label{tab:finetune_real}
\vspace{-5pt}
\end{table}

\subsection{Evaluating \ourshort with Simulator}
\begin{figure}[t]
    \centering    
    \includegraphics[width=\linewidth]{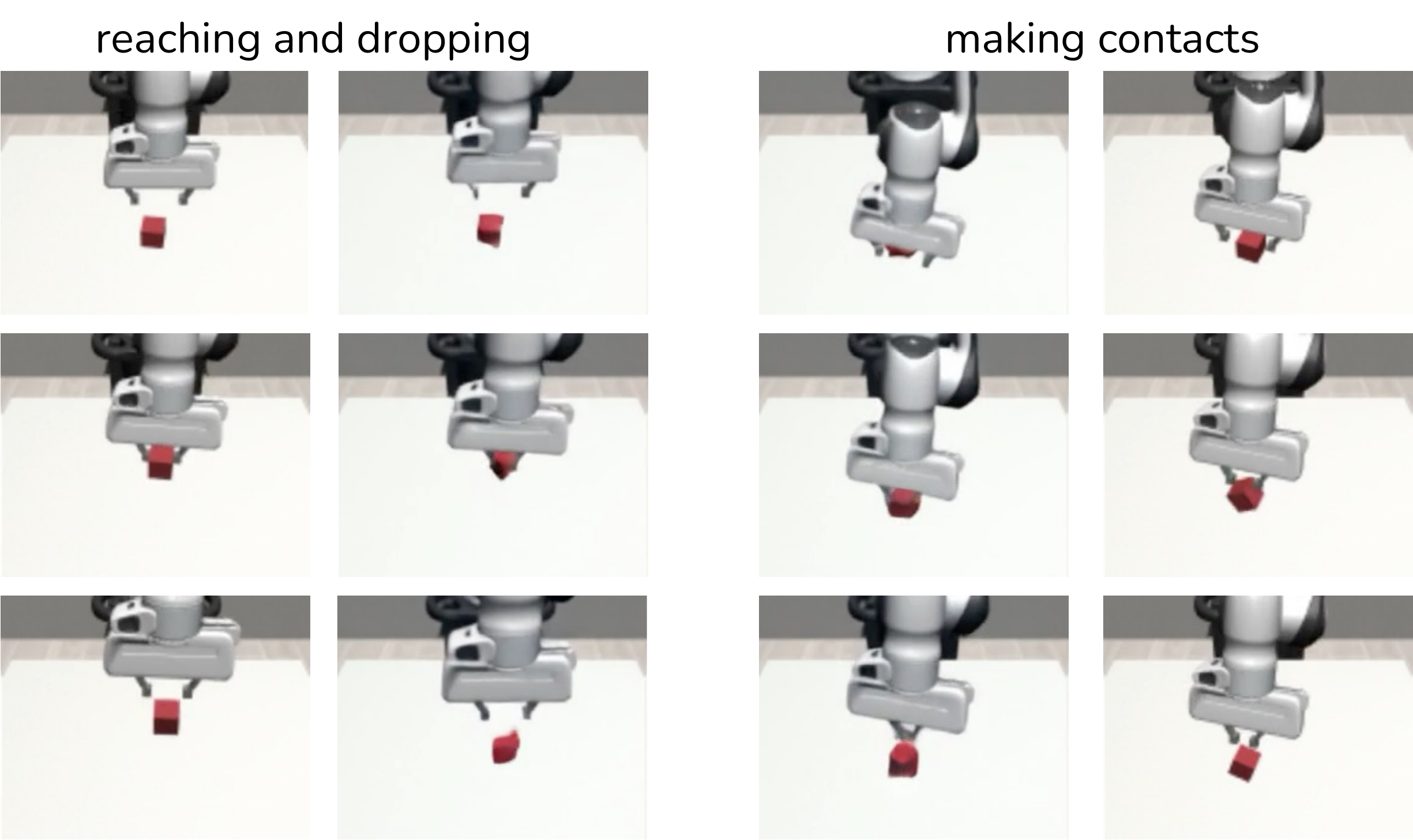}
\caption{\textbf{Policy Evaluation with \ourshort.} By learning the action-video dynamics over both successful and failed examples, \ourshort can be used to evaluate policies, similar to a traditional simulator \cite{todorov2012mujoco}. The autoregressive horizon at inference time is 10 times more than the training time horizon. }
    \label{fig:sim_app}
\end{figure}

In this section, we evaluate the controllability and visual fidelity of \ourshort against the ground truth in simulation. In \cref{table:sim}, we observe that pre-trained models can help with the performance compared to training from scratch. Moreover, through the PSNR$^*$   and Perplexity$^*$ metrics with ground truth, we observe better robustness and less video divergence with pre-training on large amounts of heterogeneous data.   In \cref{fig:sim_app}, we show some interaction examples with the learned video models compared to the ground truth simulation. The experiments show that the learned simulator is reactive despite being finetuned on only 200 trajectories. 

\begin{table}[t]
\centering
\small
\tablestyle{7pt}{1.2}
\begin{tabular}{l|cccc}
& 
{\bf PSNR$\uparrow$} &  {\bf Perplexity$\downarrow$} & {\bf PSNR$^*$$\uparrow$} & {\bf Perplexity$^*$$\downarrow$}    \\ 
\hline
\ourshort& 24.17  & 20.69 & 19.19 & 1193.70\\
\ourshort$^{+}$ & 25.11 &11.82 & 20.20 & 103.01 
\end{tabular}
\caption{\textbf{Simulation Transfer Learning.} We show that  pre-trained \ourshort can help with fine-tuning using cross-entropy losses and diffusion losses jointly. where \ourshort$^{+}$ denotes the finetuned model based on pre-trained checkpoints.}
\label{table:sim}
\end{table}

\subsection{\ourshort for Policy Evaluation}
\label{exp:finetune_sim_evaluation}

In this section, we show that a well-trained video model can be used as a simulator for evaluating policy performance. Note that since a policy can be flawed, a simulator must simulate both succeeded and failed examples, which might come from broader sets of interaction data than high-quality demonstrations, as shown in \cref{fig:sim_app}. Note that learned models for policy evaluation can be used with any particular policy. Specifically, a Diffusion Policy (DP) \cite{chi2023diffusion} is trained on Robomimic Lift \cite{robomimic2021} task with 200 trajectories for 10k iterations with an action horizon of 16. In \cref{table:eval_policy}, we present the evaluation results of four baseline policies, each trained with the same data but differing in training iterations before full convergence. We train \ourshort on the perturbed datasets generated earlier to evaluate groundtruth PSNR.

We evaluate each policy using 50 runs and observe a correlation of evaluation statistics between policies evaluated in the Mujoco Simulator \cite{todorov2012mujoco} and policies evaluated in the learned simulators. 
Note that since rewards and success criteria rely on the ground truth states, we use manual human annotation to determine the evaluation results using learned simulators. This can be done in batches and extending to using foundation models for reward labeling can be future works. There are still failure cases in capturing the physical details of dropping and catching the boxes. But we also see many highly realistic interactions with only limited amounts of data. Notably, even though the training horizon is only 12 steps, the autoregressive horizon of \ourshort at inference time can generalize to over 100 steps.

We also report the time taken to evaluate policies in both ground truth and learned simulators. Each episode in the simulation has a maximum of 100 timesteps and the average simulation time is 32 seconds. This is still slower compared to 9 seconds in the Mujoco simulator, but the potential of simulating deformable, granular objects in cluttered realistic scenes for policy evaluation is promising.

\begin{table}[t]
\centering
\small
\tablestyle{8pt}{1.2}
\begin{tabular}{l|ccccc|}
\textbf{Policy Evaluator} & 1 & 2 & 3 & 4  \\ 
\hline
{Ground Truth Simulator} & 0.38 & 0.52 & 0.70  & 1.00   \\ 
{\ourshort Simulator} & 0.43 & 0.56 & 0.66 & 0.73  
\end{tabular}
\caption{\textbf{Policy Evaluation Results Across 4 Different Policies.} We observed positive correlations of  the evaluation results  for 4 different policies bewteen the ground truth and learned simulators. The Pearson ratio between evaluations is 0.95.  }
\label{table:eval_policy}
\end{table}

\subsection{Synthetic Data Generation}
We investigate generating synthetic data for policy training using \ourshort. We leverage a dataset of 100 action trajectories to generate synthetic data (video, action) pairs based on a trained \ourshort. 

We evaluate the quality of the generated data by adding different amounts of them to 10 original trajectories. We then train baseline DP policies for 10k iterations and evaluate for 50 trials.
In \cref{table:synthetic_data}, we show policy improvements by training on 90\% of synthetic data in the low-data regime, to the extent that matches the full original data with 100 trajectories. This also leverages \ourshort's ability to generate long-horizon videos without much degradation.  

We conduct similar experiments in the Language Table \cite{lynch2023interactive} benchmark. Due to no access to real robots, we measure the data quality 
by validation learning losses. We use 10 out of 100 
trajectories from the original benchmark. We then replay and render-augment it with \ourshort from 10 to 90 trajectories with image observations generated by our systems. The training pipeline has language and image inputs and we measure the MSE losses. Using a small diffusion policies (17M), we show an improved BC policy performance trained with generated data from HMA.

We plan to explore removing the dependence on action trajectories and initial states. We can also scale up training and generating more unseen trajectories in complex real-world settings. We explored using dynamics models for policies but did not achieve surprising results. The hypothesis is that the perception used to 
predict actions is still not unified with the image generation 
objectives with VQ codes.  One underexplored capability   of autoregression in robotic policies is that a single frame can allow autoregressive generation of longer action sequences beyond the training context, in contrast to full sequence generation policies such as ACT \cite{zhao2023learning} and DP \cite{chi2023diffusion}.

\begin{table}[t]
\centering
\small
\tablestyle{4pt}{1.2}
\begin{tabular}{l|ccccc}
& +0 & +10 & +50  & +90 & original \\ 
\hline
{Success in \cite{robomimic2021}} & 82\% &  90\% & 96\%& 100\% & 100\%   \\
{Validation Loss in \cite{lynch2023interactive}} &  1.72  & 1.16   & 1.09  &  0.88 & 0.87 
\end{tabular}
\caption{\textbf{Synthetic Data for Policy Learning.} We evaluate the quality of generated synthetic data by adding different numbers of generated video trajectories in \cite{robomimic2021} and \cite{lynch2023interactive}, from 10 to 100, to a fixed subset (10 trajectories) of the original data (100 trajectories). We then conduct policy training and evaluation and report the Robomimic success rates (top row) and Language Table validation losses (bottom row).  }
\label{table:synthetic_data}
\end{table}

\section{Conclusion}
\label{sec:conclusion}
In this work, we present \ourshort, a masked autoregression approach for training action-video dynamics models from heterogeneous data. Towards this direction, we investigate the scaling behavior of video models in heterogeneous pre-training. We explore multiple downstream applications in robotics and demonstrate the generality and efficiency of the framework. We demonstrate that real-time interactive video simulation, much faster than previous works, can be learned. \ourshort is also used to evaluate policy training, generate synthetic data, and act as policies.

Limitations include the imperfect controllability of the dynamic models with limited data as well as the limited policy performance for action generation. Future works include studying autoregressive policy performance with real robots and complex setups, generating synthetic data at a large scale, and investigating world models for long-horizon planning and model predictive control. We hope this work will shed some light on building interactive world models for embodied intelligence. 

\section{Acknowledgements}
\label{sec:acknowledgement}
We would like to thank Kaiming He, Tianhong Li, Shuang Li, Yilun Du, and Xiaolong Wang  for their early discussions and suggestions. 
We also thank 1X for open-sourcing the 1XGPT starter code and tokenizer.

{\small
\bibliographystyle{ieee_fullname}
\bibliography{egbib}

\begin{thebibliography}{10}\itemsep=-1pt

\bibitem{1X_Technologies_1X_World_Model_2024}
{1X Technologies}.
\newblock {1X World Model Challenge}, June 2024.

\bibitem{alonso2024diffusion}
Eloi Alonso, Adam Jelley, Vincent Micheli, Anssi Kanervisto, Amos Storkey, Tim Pearce, and Fran{\c{c}}ois Fleuret.
\newblock Diffusion for world modeling: Visual details matter in atari.
\newblock {\em arXiv preprint arXiv:2405.12399}, 2024.

\bibitem{bertsekas1995neuro}
Dimitri~P Bertsekas and John~N Tsitsiklis.
\newblock Neuro-dynamic programming: an overview.
\newblock In {\em Proceedings of 1995 34th IEEE conference on decision and control}, volume~1, pages 560--564. IEEE, 1995.

\bibitem{blattmann2023stable}
Andreas Blattmann, Tim Dockhorn, Sumith Kulal, Daniel Mendelevitch, Maciej Kilian, Dominik Lorenz, Yam Levi, Zion English, Vikram Voleti, Adam Letts, et~al.
\newblock Stable video diffusion: Scaling latent video diffusion models to large datasets.
\newblock {\em arXiv preprint arXiv:2311.15127}, 2023.

\bibitem{videoworldsimulators2024}
Tim Brooks, Bill Peebles, Connor Holmes, Will DePue, Yufei Guo, Li Jing, David Schnurr, Joe Taylor, Troy Luhman, Eric Luhman, Clarence Ng, Ricky Wang, and Aditya Ramesh.
\newblock Video generation models as world simulators.
\newblock 2024.

\bibitem{bruce2024genie}
Jake Bruce, Michael~D Dennis, Ashley Edwards, Jack Parker-Holder, Yuge Shi, Edward Hughes, Matthew Lai, Aditi Mavalankar, Richie Steigerwald, Chris Apps, et~al.
\newblock Genie: Generative interactive environments.
\newblock In {\em Forty-first International Conference on Machine Learning}, 2024.

\bibitem{byravan2017se3}
Arunkumar Byravan and Dieter Fox.
\newblock Se3-nets: Learning rigid body motion using deep neural networks.
\newblock In {\em 2017 IEEE International Conference on Robotics and Automation (ICRA)}, pages 173--180. IEEE, 2017.

\bibitem{chang2022maskgit}
Huiwen Chang, Han Zhang, Lu Jiang, Ce Liu, and William~T Freeman.
\newblock Maskgit: Masked generative image transformer.
\newblock In {\em Proceedings of the IEEE/CVF Conference on Computer Vision and Pattern Recognition}, pages 11315--11325, 2022.

\bibitem{chen2021decision}
Lili Chen, Kevin Lu, Aravind Rajeswaran, Kimin Lee, Aditya Grover, Misha Laskin, Pieter Abbeel, Aravind Srinivas, and Igor Mordatch.
\newblock Decision transformer: Reinforcement learning via sequence modeling.
\newblock {\em Advances in neural information processing systems}, 34:15084--15097, 2021.

\bibitem{chen2023genaug}
Zoey Chen, Sho Kiami, Abhishek Gupta, and Vikash Kumar.
\newblock Genaug: Retargeting behaviors to unseen situations via generative augmentation.
\newblock {\em arXiv preprint arXiv:2302.06671}, 2023.

\bibitem{chi2023diffusion}
Cheng Chi, Zhenjia Xu, Siyuan Feng, Eric Cousineau, Yilun Du, Benjamin Burchfiel, Russ Tedrake, and Shuran Song.
\newblock Diffusion policy: Visuomotor policy learning via action diffusion.
\newblock {\em The International Journal of Robotics Research}, page 02783649241273668, 2023.

\bibitem{damen2020epic}
Dima Damen, Hazel Doughty, Giovanni~Maria Farinella, Sanja Fidler, Antonino Furnari, Evangelos Kazakos, Davide Moltisanti, Jonathan Munro, Toby Perrett, Will Price, et~al.
\newblock The epic-kitchens dataset: Collection, challenges and baselines.
\newblock {\em IEEE Transactions on Pattern Analysis and Machine Intelligence}, 43(11):4125--4141, 2020.

\bibitem{damen2022rescaling}
Dima Damen, Hazel Doughty, Giovanni~Maria Farinella, Antonino Furnari, Evangelos Kazakos, Jian Ma, Davide Moltisanti, Jonathan Munro, Toby Perrett, Will Price, et~al.
\newblock Rescaling egocentric vision: Collection, pipeline and challenges for epic-kitchens-100.
\newblock {\em International Journal of Computer Vision}, pages 1--23, 2022.

\bibitem{doshi2024scaling}
Ria Doshi, Homer Walke, Oier Mees, Sudeep Dasari, and Sergey Levine.
\newblock Scaling cross-embodied learning: One policy for manipulation, navigation, locomotion and aviation.
\newblock {\em arXiv preprint arXiv:2408.11812}, 2024.

\bibitem{du2023video}
Yilun Du, Mengjiao Yang, Pete Florence, Fei Xia, Ayzaan Wahid, Brian Ichter, Pierre Sermanet, Tianhe Yu, Pieter Abbeel, Joshua~B Tenenbaum, et~al.
\newblock Video language planning.
\newblock {\em arXiv preprint arXiv:2310.10625}, 2023.

\bibitem{finn2017deep}
Chelsea Finn and Sergey Levine.
\newblock Deep visual foresight for planning robot motion.
\newblock In {\em 2017 IEEE International Conference on Robotics and Automation (ICRA)}, pages 2786--2793. IEEE, 2017.

\bibitem{grauman2024ego}
Kristen Grauman, Andrew Westbury, Lorenzo Torresani, Kris Kitani, Jitendra Malik, Triantafyllos Afouras, Kumar Ashutosh, Vijay Baiyya, Siddhant Bansal, Bikram Boote, et~al.
\newblock Ego-exo4d: Understanding skilled human activity from first-and third-person perspectives.
\newblock In {\em Proceedings of the IEEE/CVF Conference on Computer Vision and Pattern Recognition}, pages 19383--19400, 2024.

\bibitem{ha2018world}
David Ha and J{\"u}rgen Schmidhuber.
\newblock World models.
\newblock {\em arXiv preprint arXiv:1803.10122}, 2018.

\bibitem{hansen2023td}
Nicklas Hansen, Hao Su, and Xiaolong Wang.
\newblock Td-mpc2: Scalable, robust world models for continuous control.
\newblock {\em arXiv preprint arXiv:2310.16828}, 2023.

\bibitem{he2022latent}
Yingqing He, Tianyu Yang, Yong Zhang, Ying Shan, and Qifeng Chen.
\newblock Latent video diffusion models for high-fidelity long video generation.
\newblock {\em arXiv preprint arXiv:2211.13221}, 2022.

\bibitem{heusel2017gans}
Martin Heusel, Hubert Ramsauer, Thomas Unterthiner, Bernhard Nessler, and Sepp Hochreiter.
\newblock Gans trained by a two time-scale update rule converge to a local nash equilibrium.
\newblock {\em Advances in neural information processing systems}, 30, 2017.

\bibitem{ho2022video}
Jonathan Ho, Tim Salimans, Alexey Gritsenko, William Chan, Mohammad Norouzi, and David~J Fleet.
\newblock Video diffusion models.
\newblock {\em Advances in Neural Information Processing Systems}, 35:8633--8646, 2022.

\bibitem{hore2010image}
Alain Hore and Djemel Ziou.
\newblock Image quality metrics: Psnr vs. ssim.
\newblock In {\em 2010 20th international conference on pattern recognition}, pages 2366--2369. IEEE, 2010.

\bibitem{janner2022planning}
Michael Janner, Yilun Du, Joshua~B Tenenbaum, and Sergey Levine.
\newblock Planning with diffusion for flexible behavior synthesis.
\newblock {\em arXiv preprint arXiv:2205.09991}, 2022.

\bibitem{kirillov2023segment}
Alexander Kirillov, Eric Mintun, Nikhila Ravi, Hanzi Mao, Chloe Rolland, Laura Gustafson, Tete Xiao, Spencer Whitehead, Alexander~C Berg, Wan-Yen Lo, et~al.
\newblock Segment anything.
\newblock {\em arXiv preprint arXiv:2304.02643}, 2023.

\bibitem{kondratyuk2023videopoet}
Dan Kondratyuk, Lijun Yu, Xiuye Gu, Jos{\'e} Lezama, Jonathan Huang, Grant Schindler, Rachel Hornung, Vighnesh Birodkar, Jimmy Yan, Ming-Chang Chiu, et~al.
\newblock Videopoet: A large language model for zero-shot video generation.
\newblock {\em arXiv preprint arXiv:2312.14125}, 2023.

\bibitem{li2024autoregressive}
Tianhong Li, Yonglong Tian, He Li, Mingyang Deng, and Kaiming He.
\newblock Autoregressive image generation without vector quantization.
\newblock {\em arXiv preprint arXiv:2406.11838}, 2024.

\bibitem{li2018learning}
Yunzhu Li, Jiajun Wu, Russ Tedrake, Joshua~B Tenenbaum, and Antonio Torralba.
\newblock Learning particle dynamics for manipulating rigid bodies, deformable objects, and fluids.
\newblock {\em arXiv preprint arXiv:1810.01566}, 2018.

\bibitem{liu2024mardini}
Haozhe Liu, Shikun Liu, Zijian Zhou, Mengmeng Xu, Yanping Xie, Xiao Han, Juan~C P{\'e}rez, Ding Liu, Kumara Kahatapitiya, Menglin Jia, et~al.
\newblock Mardini: Masked autoregressive diffusion for video generation at scale.
\newblock {\em arXiv preprint arXiv:2410.20280}, 2024.

\bibitem{luo2024open}
Zhuoyan Luo, Fengyuan Shi, Yixiao Ge, Yujiu Yang, Limin Wang, and Ying Shan.
\newblock Open-magvit2: An open-source project toward democratizing auto-regressive visual generation.
\newblock {\em arXiv preprint arXiv:2409.04410}, 2024.

\bibitem{lynch2023interactive}
Corey Lynch, Ayzaan Wahid, Jonathan Tompson, Tianli Ding, James Betker, Robert Baruch, Travis Armstrong, and Pete Florence.
\newblock Interactive language: Talking to robots in real time.
\newblock {\em IEEE Robotics and Automation Letters}, 2023.

\bibitem{robomimic2021}
Ajay Mandlekar, Danfei Xu, Josiah Wong, Soroush Nasiriany, Chen Wang, Rohun Kulkarni, Li Fei-Fei, Silvio Savarese, Yuke Zhu, and Roberto Mart\'{i}n-Mart\'{i}n.
\newblock What matters in learning from offline human demonstrations for robot manipulation.
\newblock In {\em arXiv preprint arXiv:2108.03298}, 2021.

\bibitem{murray2017mathematical}
Richard~M Murray, Zexiang Li, and S~Shankar Sastry.
\newblock {\em A mathematical introduction to robotic manipulation}.
\newblock CRC press, 2017.

\bibitem{nair2018visual}
Ashvin~V Nair, Vitchyr Pong, Murtaza Dalal, Shikhar Bahl, Steven Lin, and Sergey Levine.
\newblock Visual reinforcement learning with imagined goals.
\newblock {\em Advances in neural information processing systems}, 31, 2018.

\bibitem{o2023open}
Abby O'Neill, Abdul Rehman, Abhinav Gupta, Abhiram Maddukuri, Abhishek Gupta, Abhishek Padalkar, Abraham Lee, Acorn Pooley, Agrim Gupta, Ajay Mandlekar, et~al.
\newblock Open x-embodiment: Robotic learning datasets and rt-x models.
\newblock {\em arXiv preprint arXiv:2310.08864}, 2023.

\bibitem{openai2023gpt4}
OpenAI.
\newblock Gpt-4 technical report, 2023.

\bibitem{peebles2023scalable}
William Peebles and Saining Xie.
\newblock Scalable diffusion models with transformers.
\newblock In {\em Proceedings of the IEEE/CVF International Conference on Computer Vision}, pages 4195--4205, 2023.

\bibitem{radford2019language}
Alec Radford, Jeffrey Wu, Rewon Child, David Luan, Dario Amodei, Ilya Sutskever, et~al.
\newblock Language models are unsupervised multitask learners.
\newblock {\em OpenAI blog}, 1(8):9, 2019.

\bibitem{radosavovic2024humanoid}
Ilija Radosavovic, Bike Zhang, Baifeng Shi, Jathushan Rajasegaran, Sarthak Kamat, Trevor Darrell, Koushil Sreenath, and Jitendra Malik.
\newblock Humanoid locomotion as next token prediction.
\newblock {\em arXiv preprint arXiv:2402.19469}, 2024.

\bibitem{rigter2024avid}
Marc Rigter, Tarun Gupta, Agrin Hilmkil, and Chao Ma.
\newblock Avid: Adapting video diffusion models to world models.
\newblock {\em arXiv preprint arXiv:2410.12822}, 2024.

\bibitem{seo2023masked}
Younggyo Seo, Danijar Hafner, Hao Liu, Fangchen Liu, Stephen James, Kimin Lee, and Pieter Abbeel.
\newblock Masked world models for visual control.
\newblock In {\em Conference on Robot Learning}, pages 1332--1344. PMLR, 2023.

\bibitem{shazeer2017outrageously}
Noam Shazeer, Azalia Mirhoseini, Krzysztof Maziarz, Andy Davis, Quoc Le, Geoffrey Hinton, and Jeff Dean.
\newblock Outrageously large neural networks: The sparsely-gated mixture-of-experts layer.
\newblock {\em arXiv preprint arXiv:1701.06538}, 2017.

\bibitem{song2020denoising}
Jiaming Song, Chenlin Meng, and Stefano Ermon.
\newblock Denoising diffusion implicit models.
\newblock {\em arXiv preprint arXiv:2010.02502}, 2020.

\bibitem{team2024octo}
Octo~Model Team, Dibya Ghosh, Homer Walke, Karl Pertsch, Kevin Black, Oier Mees, Sudeep Dasari, Joey Hejna, Tobias Kreiman, Charles Xu, et~al.
\newblock Octo: An open-source generalist robot policy.
\newblock {\em arXiv preprint arXiv:2405.12213}, 2024.

\bibitem{tian2024visual}
Keyu Tian, Yi Jiang, Zehuan Yuan, Bingyue Peng, and Liwei Wang.
\newblock Visual autoregressive modeling: Scalable image generation via next-scale prediction.
\newblock {\em arXiv preprint arXiv:2404.02905}, 2024.

\bibitem{todorov2012mujoco}
Emanuel Todorov, Tom Erez, and Yuval Tassa.
\newblock Mujoco: A physics engine for model-based control.
\newblock In {\em 2012 IEEE/RSJ international conference on intelligent robots and systems}, pages 5026--5033. IEEE, 2012.

\bibitem{unterthiner2018towards}
Thomas Unterthiner, Sjoerd Van~Steenkiste, Karol Kurach, Raphael Marinier, Marcin Michalski, and Sylvain Gelly.
\newblock Towards accurate generative models of video: A new metric \& challenges.
\newblock {\em arXiv preprint arXiv:1812.01717}, 2018.

\bibitem{valevski2024diffusionmodelsrealtimegame}
Dani Valevski, Yaniv Leviathan, Moab Arar, and Shlomi Fruchter.
\newblock Diffusion models are real-time game engines, 2024.

\bibitem{wang2024scaling}
Lirui Wang, Xinlei Chen, Jialiang Zhao, and Kaiming He.
\newblock Scaling proprioceptive-visual learning with heterogeneous pre-trained transformers.
\newblock {\em arXiv preprint arXiv:2409.20537}, 2024.

\bibitem{wang2004image}
Zhou Wang, Alan~C Bovik, Hamid~R Sheikh, and Eero~P Simoncelli.
\newblock Image quality assessment: from error visibility to structural similarity.
\newblock {\em IEEE transactions on image processing}, 13(4):600--612, 2004.

\bibitem{yang2022tensor}
Greg Yang, Edward~J Hu, Igor Babuschkin, Szymon Sidor, Xiaodong Liu, David Farhi, Nick Ryder, Jakub Pachocki, Weizhu Chen, and Jianfeng Gao.
\newblock Tensor programs v: Tuning large neural networks via zero-shot hyperparameter transfer.
\newblock {\em arXiv preprint arXiv:2203.03466}, 2022.

\bibitem{yang2023learning}
Mengjiao Yang, Yilun Du, Kamyar Ghasemipour, Jonathan Tompson, Dale Schuurmans, and Pieter Abbeel.
\newblock Learning interactive real-world simulators.
\newblock {\em arXiv preprint arXiv:2310.06114}, 2023.

\bibitem{ye2024latent}
Seonghyeon Ye, Joel Jang, Byeongguk Jeon, Sejune Joo, Jianwei Yang, Baolin Peng, Ajay Mandlekar, Reuben Tan, Yu-Wei Chao, Bill~Yuchen Lin, et~al.
\newblock Latent action pretraining from videos.
\newblock {\em arXiv preprint arXiv:2410.11758}, 2024.

\bibitem{yu2020meta}
Tianhe Yu, Deirdre Quillen, Zhanpeng He, Ryan Julian, Karol Hausman, Chelsea Finn, and Sergey Levine.
\newblock Meta-world: A benchmark and evaluation for multi-task and meta reinforcement learning.
\newblock In {\em Conference on robot learning}, pages 1094--1100. PMLR, 2020.

\bibitem{yu2023scaling}
Tianhe Yu, Ted Xiao, Austin Stone, Jonathan Tompson, Anthony Brohan, Su Wang, Jaspiar Singh, Clayton Tan, Jodilyn Peralta, Brian Ichter, et~al.
\newblock Scaling robot learning with semantically imagined experience.
\newblock {\em arXiv preprint arXiv:2302.11550}, 2023.

\bibitem{zhang2021autoregressive}
Michael~R Zhang, Tom~Le Paine, Ofir Nachum, Cosmin Paduraru, George Tucker, Ziyu Wang, and Mohammad Norouzi.
\newblock Autoregressive dynamics models for offline policy evaluation and optimization.
\newblock {\em arXiv preprint arXiv:2104.13877}, 2021.

\bibitem{zhang2018perceptual}
Richard Zhang, Phillip Isola, Alexei~A Efros, Eli Shechtman, and Oliver Wang.
\newblock The unreasonable effectiveness of deep features as a perceptual metric.
\newblock In {\em CVPR}, 2018.

\bibitem{zhao2023learning}
Tony~Z Zhao, Vikash Kumar, Sergey Levine, and Chelsea Finn.
\newblock Learning fine-grained bimanual manipulation with low-cost hardware.
\newblock {\em arXiv preprint arXiv:2304.13705}, 2023.

\bibitem{zhou2023nerf}
Allan Zhou, Moo~Jin Kim, Lirui Wang, Pete Florence, and Chelsea Finn.
\newblock Nerf in the palm of your hand: Corrective augmentation for robotics via novel-view synthesis.
\newblock In {\em Proceedings of the IEEE/CVF Conference on Computer Vision and Pattern Recognition}, pages 17907--17917, 2023.

\bibitem{zhu2024irasim}
Fangqi Zhu, Hongtao Wu, Song Guo, Yuxiao Liu, Chilam Cheang, and Tao Kong.
\newblock Irasim: Learning interactive real-robot action simulators.
\newblock {\em arXiv preprint arXiv:2406.14540}, 2024.

\end{thebibliography}
}

\end{document}